\documentclass{article}
\usepackage{standalone}
\usepackage{pgfplots}

\usepackage[numbers]{natbib}
\usepackage[final]{template/nips_2017}

\usepackage[utf8]{inputenc} 
\usepackage[T1]{fontenc}     
\usepackage{hyperref}          
\usepackage{url}                   
\usepackage{booktabs}         
\usepackage{amsfonts}         
\usepackage{nicefrac}           
\usepackage{microtype}        

\usepackage[cmex10]{amsmath}
\usepackage{amsfonts}
\usepackage{bm}

\usepackage{lmodern}

\usepackage{physics}

\usepackage[section]{placeins}
\usepackage{geometry,array,graphicx,float}
\usepackage{subcaption}
\usepackage{enumitem}
\usepackage[font=footnotesize]{caption}

\title{Pre-trainable Reservoir Computing with Recursive Neural Gas}

\author{Luca Carcano \\
	Computer Vision and Multimedia Lab\\
  	Universit\`{a} Degli Studi Di Pavia \\
  \texttt{luca.carcano@gmail.com} \\
  \AND
  Emanuele Plebani \\ 
  STMicroelectronics \\ 
  \texttt{emanuele.plebani1@st.com} \\
  \And
  Danilo Pietro Pau \\ 
  STMicroelectronics \\
  \texttt{danilo.pau@st.com} \\
  \And
  Marco Piastra \\
  Computer Vision and Multimedia Lab \\ 
  Universit\`{a} Degli Studi Di Pavia \\
  \texttt{marco.piastra@unipv.it}
}

\begin{document}

\maketitle

\begin{abstract}
Echo State Networks (ESN) are a class of Recurrent Neural Networks (RNN) that has gained substantial popularity due to their effectiveness, ease of use and potential for compact hardware implementation. An ESN contains the three network layers \emph{input}, \emph{reservoir} and \emph{readout} where the reservoir is the truly recurrent network. The input and reservoir layers of an ESN are initialized at random and never trained afterwards and the training of the ESN is applied to the readout layer only. The alternative of Recursive Neural Gas (RNG) is one of the many proposals of fully-trainable reservoirs that can be found in the literature. Although some improvements in performance have been reported with RNG, to the best of authors' knowledge, no experimental comparative results are known with benchmarks for which ESN is known to yield excellent results. This work describes an accurate model of RNG together with some extensions to the models presented in the literature and shows comparative results on three well-known and accepted datasets. The experimental results obtained show that, under specific circumstances, RNG-based reservoirs can achieve better performance.
\end{abstract}

\section{Introduction}


Reservoir computing is a computational framework for recurrent neural networks in which the input signal is fed into a large, recurrent pool of neurons called reservoir. The reservoir is used to map the input to a higher dimension and a simple readout layer (usually a linear or ridge regression) is then trained to read the state of the reservoir and map to the desired output. Notable examples of reservoir computing systems are Liquid State Machines \citep{Maass2010} and Echo State Networks (ESN) \citep{Jaeger2002}.

Based on the model proposed in \citep{Jaeger2002} for Echo State Networks, we consider here a reservoir computing system made up of three distinct layers: 
\begin{figure}[h]
  \centering
  \includegraphics[width=.4\linewidth]{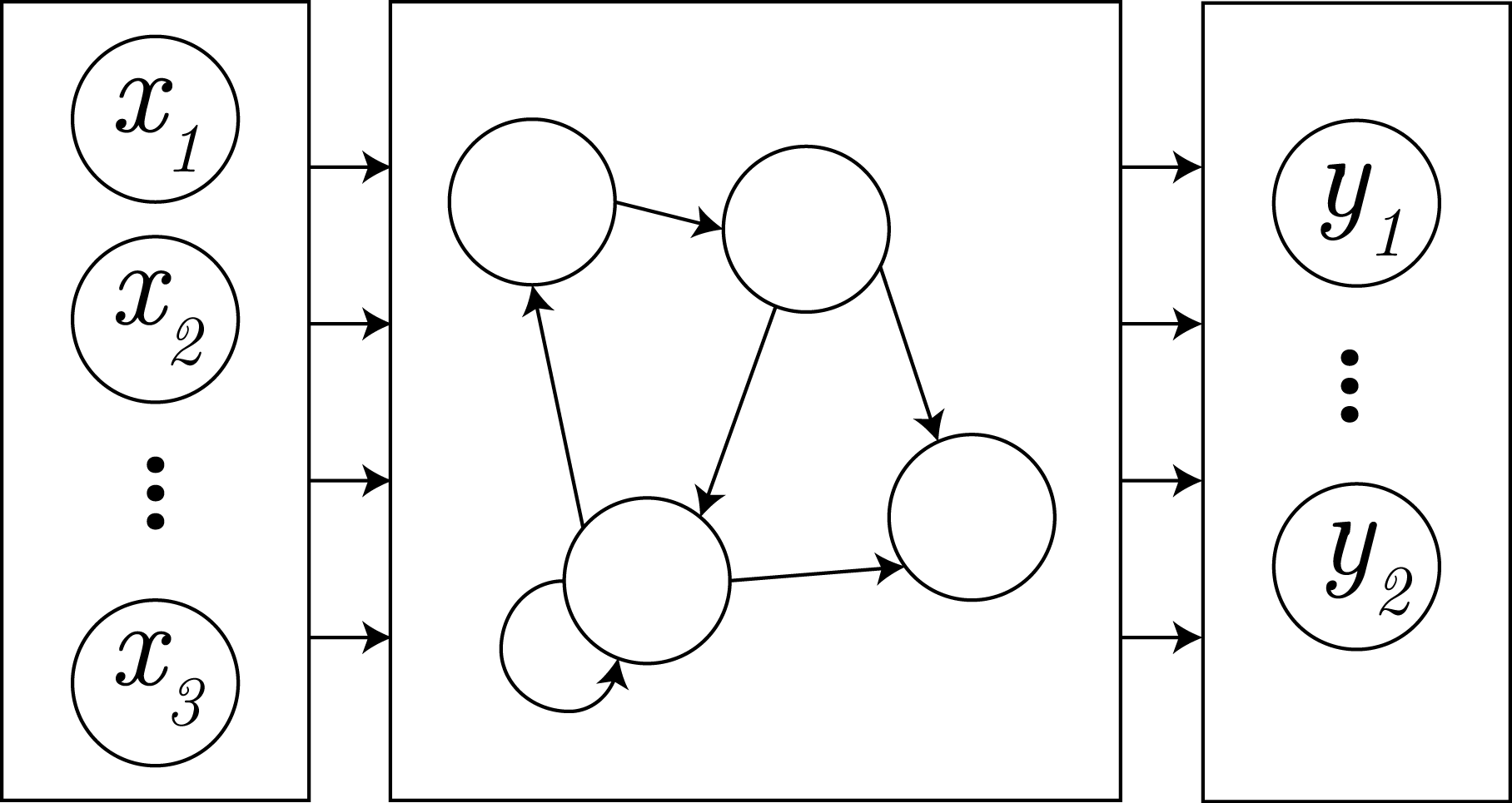}
\end{figure}
\begin{enumerate}
	\item \emph{input layer}, which maps the input signal onto the reservoir in feed-forward mode; 
	\item \emph{reservoir layer}, which is the truly recursive neural net (RNN).
	\item \emph{readout layer}, which is a feed-forward neural network that maps the state of the reservoir to the output desired.
\end{enumerate}

With ESNs, the readout layer is the only component of the network that is trained, via supervised training, while the weights in both the input and the reservoir are initialized at random, with some post-processing (see below), and never trained afterwards \citep{Jaeger2002}. Nonetheless, ESNs have reached a substantial popularity in the field due to the ease of their implementation, including the possibility of being realized in hardware \citep{Paquot2012,Vandoorne2014}.

On the other hand, in the light of the success of other machine learning models, many proposal have been made to introduce some form of unsupervised pre-training for the input and reservoir layers as well. Recursive Neural Gas (RNG) \citep{Voegtlin2002}, in particular has been proposed as a model in \citep{Gallicchio2011, Lukosevicius2012, Boccato2014}. Although some improvements in performance have been reported with RNG, to the best of our knowledge, no experimental comparative results are known with benchmarks for which ESN is known to yield excellent results. This work describes first an accurate model of RNG together with some extensions to the models presented in the literature and then shows comparative results on three well-known benchmark datasets. The experimental results obtained show that, under specific circumstances to be described, RNG-based reservoirs can indeed achieve better performance. 

\section{A Pre-trainable Reservoir Model}
\label{section_2}

\subsection{Neural Gas}

Given an input data distribution described by probability $P(\bm{x})$, where $\bm{x} \in \mathbb{R}^d$, a \emph{neural gas} (NG) \citep{Martinetz1993} is a set $U$ of $n$ \emph{units}, each associated to a reference vector in $\mathbb{R}^d$:
$$U := \{\bm{w}_i\},\ \bm{w}_i \in \mathbb{R}^d,\ i \in \{1,\ldots,n\}.$$
The unsupervised training of the NG occurs by adapting the reference vectors in $U$ to the input probability $P$ by repeating the following iteration:
\begin{enumerate} [noitemsep,topsep=0pt,parsep=0pt,partopsep=0pt]
  \item receive one signal $\bm{x}$ distributed as $P(\bm{x})$;
  \item update the reference vectors in $U$ (see below);
  \item return to step 1.
\end{enumerate}
In each iteration, the reference vectors in $U$ are updated by
\begin{equation}
\label{eq:ng_update}
\Delta \bm{w}_{i} = \varepsilon\cdot h_{\lambda}(k_i(\bm{x})) \cdot (\bm{x} - \bm{w}_{i})
\end{equation}
where $k_i(\bm{x}) := \#\{\bm{w}_{j} : | \bm{x} - \bm{w}_{j} | < | \bm{x} - \bm{w}_{i} |\}$ ($\#$ denotes the cardinality), $\varepsilon > 0$ is a real parameter, 
$h_{0}(k) := \delta_{0k}
 \quad \text{and} \quad h_{\lambda}(k) := e^{-\frac{k}{\lambda}}, \text{ for } \lambda > 0$
(throughout this paper, $\delta$ denotes the usual Kronecker delta function).\\
Note that For $\lambda \rightarrow 0$, equation \eqref{eq:ng_update} becomes equivalent to
\begin{equation}
\label{eq:km_update}
\Delta \bm{w}_{i} = \varepsilon \cdot \delta_{ic_1(\bm{x})} \cdot (\bm{x} - \bm{w}_{i})
\end{equation}
where $c_1(\bm{x})$ is the function that returns the index of the closest neighbor in $U$ to the input signal $\bm{x}$. Equation \eqref{eq:km_update} is the update law of the well-known K-means algorithm \citep{Lloyd1982, Macqueen1967}.

\subsubsection{Convergence}

It is proven in \citep{Martinetz1993} that the NG algorithm performs a \emph{stochastic gradient descent} (SGD) over the energy function:
\begin{equation}
\label{eq:ng_energy}
E_{NG}(U) = \frac{1}{2C_{\lambda}}\sum_{i=1}^n{\int_V{P(\bm{x}) h_{\lambda}(k_i(\bm{x})) (\bm{x} - \bm{w}_{i})^2 d\bm{x}}}
\end{equation}
where $V$ is the support of probability $P(\bm{x})$ and
$C_{\lambda} := \sum_{i=0}^{n-1}{h_{\lambda}(i)}.$
In particular, in \citep{Martinetz1993} it is proven that
$\frac{\partial E_{NG}}{\partial \bm{w}_{i}} = - \frac{1}{C_{\lambda}}\int_V{P(\bm{x}) h_{\lambda}(k_i(\bm{x})) (\bm{x} - \bm{w}_{i}) d\bm{x}}$
which makes equation \eqref{eq:ng_update} an SGD update law. In keeping with this, an NG can be made to converge to a steady configuration by choosing values of $\varepsilon$ that decrease exponentially with the iterations of the algorithm:
\begin{equation}
\label{eq:decaying_epsilon}
  \varepsilon(t) := \varepsilon_i\left(\frac{\varepsilon_f}{\varepsilon_i}\right)^{t/T}
\end{equation}
where $\varepsilon_i$ and $\varepsilon_f$ are the initial and final values, respectively, $T$ is the total number of iterations and $t$ is the current iteration. According to \citep{Martinetz1993}, the NG algorithm converges faster than the K-means algorithm provided that $\lambda$ decays exponentially as the execution progresses, with the law
\begin{equation}
\label{eq:decaying_lambda}
  \lambda(t) := \lambda_i\left(\frac{\lambda_f}{\lambda_i}\right)^{t/T}.
\end{equation}

\subsection{Recursive Neural Gas}

Recursive Neural Gas (RNG) was first introduced in \citep{Voegtlin2002} as a recurrent neural network (RNN) based on NG. In RNG, the set of units $U$ becomes
$$U := \{(\bm{w}^{in}_i, \bm{w}^{rec}_i)\},\ \bm{w}^{in}_i \in \mathbb{R}^d,\ \bm{w}^{rec}_i \in \mathbb{R}^n,\ i \in \{1,\ldots,n\}.$$
 
The \emph{recursive transfer function} of RNG is 
\begin{align}
\label{eq:rec_trans_func}
	\tilde{v}_{i}(t) &= \exp \left( - \alpha  \left| \bm{w}^{in}_{i} - \bm{x}(t) \right|^2 - \beta \left| \bm{w}^{rec}_{i} - \bm{v}(t-1) \right|^2 \right) \\
\label{eq:leaky_int}	
	v_i(t) &= (1 - \gamma) v_i(t-1) + \gamma \tilde{v}_i(t)
\end{align}

where $\alpha, \beta, \gamma \in \mathbb{R}^+, \gamma \le 1$. In what above, $v_i(t)$ is the \emph{state} of unit $i$ at iteration $t$ and it depends on both the input $\bm{x}(t)$ at the same iteration $t$ and the state $\bm{v}(t-1)$ of all RNG units at the previous iteration $t - 1$. In words, the overall state $\bm{v}(t)$ of RNG at iteration $t$ is computed by applying an exponential Radial Basis Function (RBF) to a weighted sum of the input and previous state, with a leaky integration model.

According to the original RNG account, the unsupervised training is performed by using the iterative method of NG with updating equations:
\begin{align}
\label{eq:update_w_in}
	\Delta \bm{w}^{in}_i &= \varepsilon \cdot \sqrt{\alpha} \cdot h_{\lambda}(k_i(\bm{x}(t))) \cdot  (\bm{x}(t) - \bm{w}^{in}_i(t-1)) \\
\label{eq:update_w_rec}
	\Delta \bm{w}^{rec}_i &= \varepsilon \cdot \sqrt{\beta} \cdot h_{\lambda}(k_i(\bm{v}(t-1))) \cdot  (\bm{v}(t-1) - \bm{w}^{rec}_i(t-1))
\end{align}

This RNG model was adopted in \citep{Lukosevicius2012} as the basis for a pre-trainable reservoir layer in a specific reservoir computing architecture.  
    
\subsection{Joint Vector Space}

One potential shortcoming of the original RNG model is that each of the two vector components of a unit $\bm{w}^{in}$ and $\bm{w}^{rec}$ is trained separately, as described by  \eqref{eq:update_w_in} and \eqref{eq:update_w_rec}, while in the recursive transfer function \eqref{eq:rec_trans_func} the two components act together. In an alternative view, each unit in $U$ could be seen as:
$$U := \{\left[\sqrt{\alpha}\,\bm{w}^{in}_i; \sqrt{\beta}\,\bm{w}^{rec}_i\right]\},\ \left[\sqrt{\alpha}\,\bm{w}^{in}_i; \sqrt{\beta}\,\bm{w}^{rec}_i\right] \in \mathbb{R}^{d + n},\ i \in \{1,\ldots,n\}.$$
where `$;$' here denotes the concatenation of the two vector components. In other words, in this  alternative view, RNG vectors are seen in a joint vector space where $\sqrt{\alpha}$ and $\sqrt{\beta}$ are the relative scale factors for combining the two components. Note that this change of perspective does not alter the recursive transfer function since
\begin{equation}
\label{eq:joint_space_norm}
\left|\left[\sqrt{\alpha}\,\bm{w}^{in}_i; \sqrt{\beta}\,\bm{w}^{rec}_i\right]\right|^2 = \alpha \left| \bm{w}^{in}_i \right|^2   +  \beta \left| \bm{w}^{rec}_i \right|^2
\end{equation}
and therefore the RBF equation \eqref{eq:rec_trans_func} remains unaffected. In contrast, the update equation for each joint vector during unsupervised training becomes:
\begin{equation}
\label{eq:joint_space_update}
\Delta \left[\sqrt{\alpha}\,\bm{w}^{in}_i; \sqrt{\beta}\,\bm{w}^{rec}_i\right] = \varepsilon  \cdot h_{\lambda}\left(k_i(\bm{s}(t)) \right) \cdot \left( \bm{s}(t) - \left[\sqrt{\alpha}\,\bm{w}^{in}_i; \sqrt{\beta}\,\bm{w}^{rec}_i\right] \right)
\end{equation}
where:
$$\bm{s}(t) := \left[\sqrt{\alpha}\,\bm{x}(t); \sqrt{\beta}\,\bm{v}(t-1)\right]$$  
and the partition function $h_{\lambda}( i, \bm{s}(t))$ is computed by using the norm \eqref{eq:joint_space_norm} defined in the joint vector space and so is the overall ranking of units.

\subsection{Masking Units}

Another aspect that is investigated here is the effect of decoupling a fraction of the units in RNG from the input signal. The motivation is that, as reported in \citep{Voegtlin2002} and \citep{Lukosevicius2012}, best performances tend to be obtained with relatively high values of the $\alpha$ parameter. From \eqref{eq:rec_trans_func}, it is evident that higher values of $\alpha$ make the recursive transfer function be more specific with respect to the input $\bm{x}(t)$ since a higher value of $\alpha$ corresponds to a shorter radius of the RBF.

In the RNG variant envisaged for this, a fraction $\eta \in [0, 1]$ of units is \emph{masked} from input, so that masked units depend on recursive states alone. For such masked units, the recursive transfer function becomes:  
\begin{equation}
\label{eq:masked_rec_trans_func}
	\tilde{v}_{i}(t) := \exp \left( - \beta^{masked} \left| \bm{w}^{rec}_{i} - \bm{v}(t-1) \right|^2 \right) 
\end{equation}

Together, the introduction of the joint vector space and unit masking lead to four distinct options for unsupervised RNG training:
\begin{enumerate}[noitemsep,topsep=0pt,parsep=0pt,partopsep=0pt]
\item keeping all units in a joint vector space;
\item having $\bm{w}^{in}$ and $\bm{w}^{rec}$ distinct but also assuming that both masked and non-masked units are kept in the same $rec$ vector space:
\item keeping $\bm{w}^{in}$ and $\bm{w}^{rec}$ in the same vector space but leaving all masked units in a separate vector space;
\item having $\bm{w}^{in}$ and $\bm{w}^{rec}$ distinct and also leaving all masked units in a separate vector space.
\end{enumerate} 

Options 1. and 2. imply that, for uniformity, $\beta = \beta^{masked}$ and, apart from using \eqref{eq:masked_rec_trans_func} for masked units only, every other aspect remains unaffected. On the other hand, options 2. and 3. imply that the following update equation is used for masked units alone:
\begin{equation}
\label{eq:masked_update}
	\Delta \bm{w}^{mask}_i := \varepsilon \cdot \sqrt{\beta^{masked}} \cdot h_{\lambda}(k_i(\bm{v}(t-1))) \cdot  (\bm{v}(t-1) - \bm{w}^{rec}_i(t-1))
\end{equation}
where, for those same units, the partition function $h_{\lambda} \left( i, \bm{v}(t-1) \right)$ is computed in the vector space of masked units. 

\subsection{RNG in Other Works }

As already mentioned, the original RNG model \citep{Voegtlin2002} is based on the transfer function \eqref{eq:rec_trans_func} and the update equations \eqref{eq:update_w_in} and \eqref{eq:update_w_rec}. In the same work, the decaying $\lambda$ technique \eqref{eq:decaying_lambda} is adopted for training. The same RNG model is adopted in \citep{Lukosevicius2012} but in that work the value of the parameter $\lambda$ is apparently kept fixed during training, i.e. $\lambda_i = \lambda_f = 1.0$. The NG model is adopted in \citep{Boccato2014} for training the input layer of an ESN, using the decaying $\lambda$ technique. The NG model is also adopted in \citep{Gallicchio2011} for training a graph-based variant of the ESN.

\begin{figure}
\centering
\begin{subfigure}[b]{0.32\textwidth}
            \includegraphics[width = \textwidth,trim={0 1cm 0 1cm},clip]{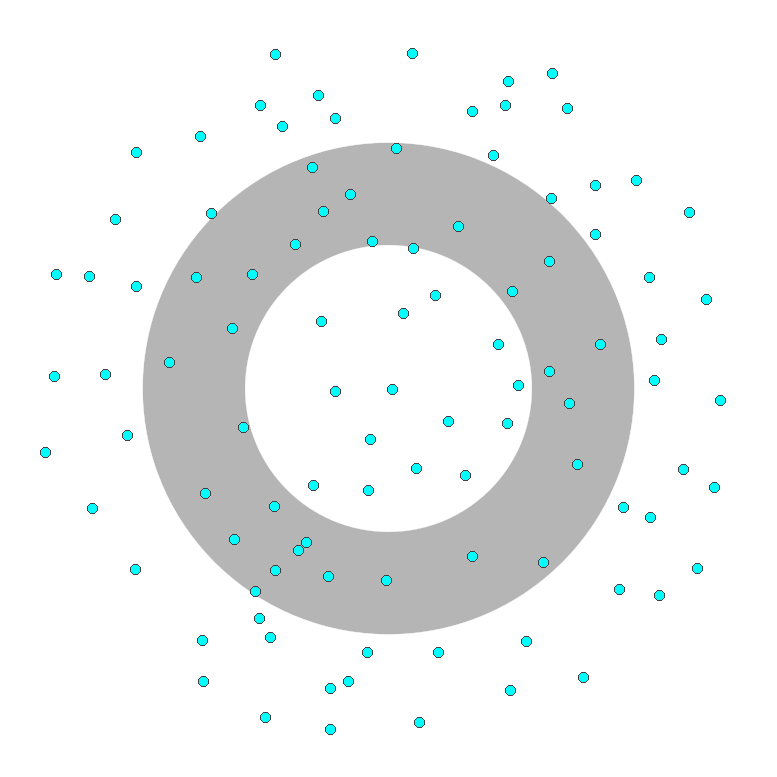}
            \caption{}
            \label{fig:plot_lambda_0}
    \end{subfigure}%
\begin{subfigure}[b]{0.32\textwidth}
            \includegraphics[width = \textwidth,trim={0 1cm 0 1cm},clip]{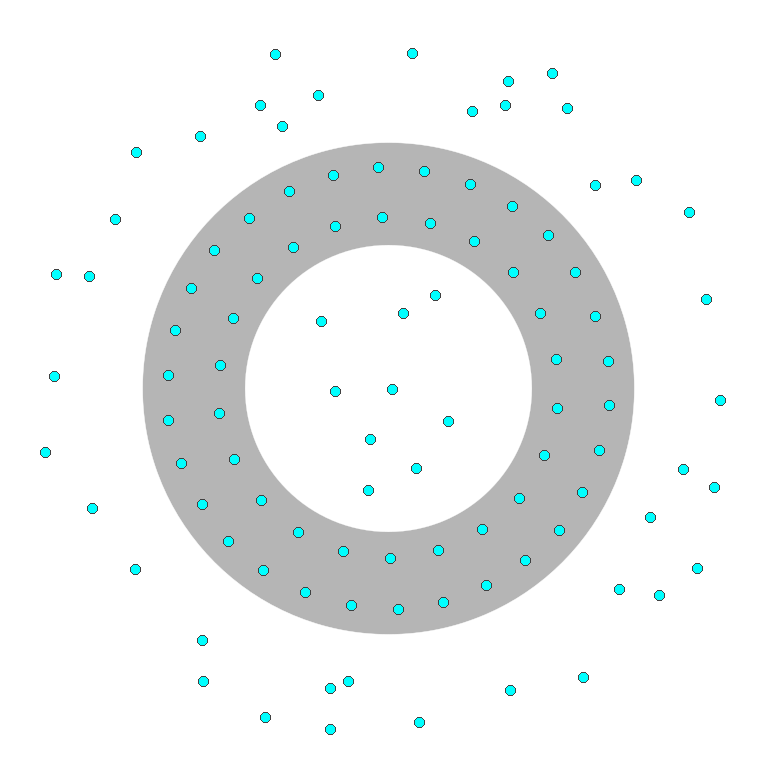}
            \caption{}
            \label{fig:plot_lambda_1}
    \end{subfigure}%
    
\begin{subfigure}[b]{0.31\textwidth}
            \includegraphics[width = \textwidth,trim={0 3cm 0 1cm},clip]{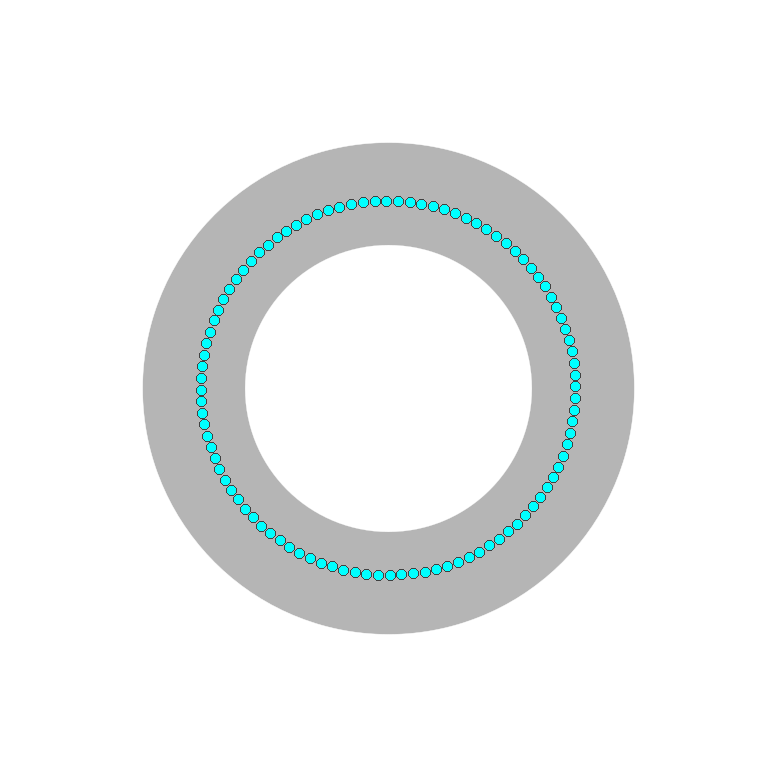}
            \caption{}
            \label{fig:plot_lambda_dec_8}
    \end{subfigure}%
    \begin{subfigure}[b]{0.31\textwidth}
            \includegraphics[width = \textwidth,trim={0 3cm 0 1cm},clip]{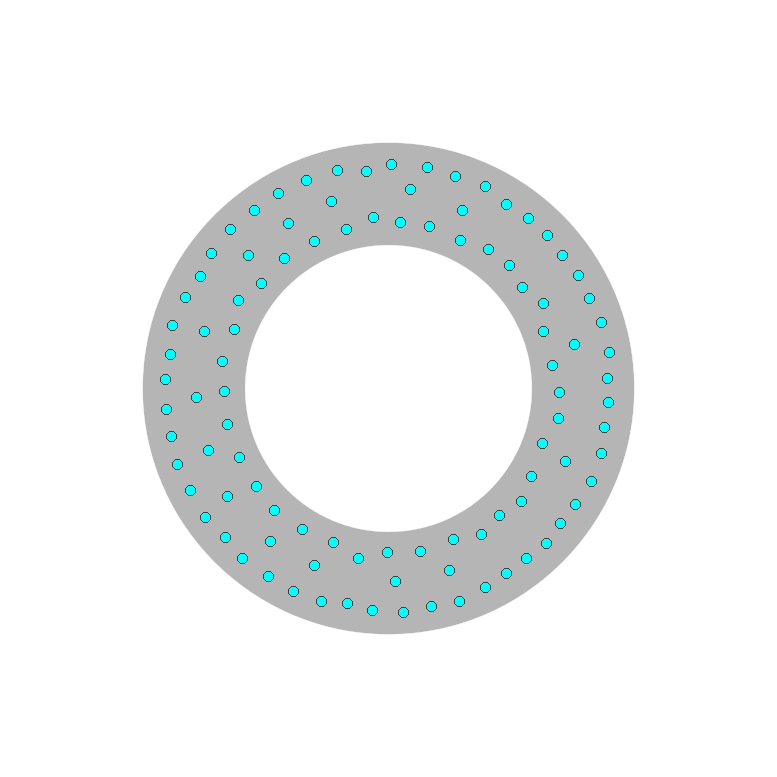}
            \caption{}
            \label{fig:plot_lambda_dec_1}
    \end{subfigure}%
    \begin{subfigure}[b]{0.31\textwidth}
            \includegraphics[width = \textwidth,trim={0 3cm 0 1cm},clip]{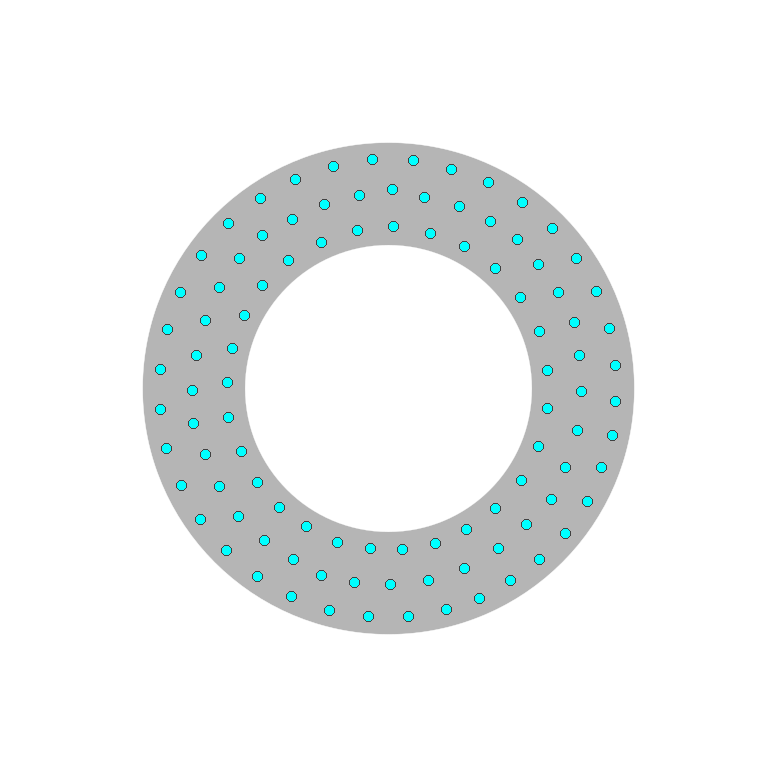}
            \caption{}
            \label{fig:plot_lambda_dec01}
    \end{subfigure}%
    \caption{\label{fig:decaying_lambda}
Behavior of 100 NG units with a ring-shaped input dataset (in grey). Initially, units are positioned uniformly at random (a) over a disc shape that includes the dataset. After 100K iterations with $\lambda_i = \lambda_f \approx 0$ (b), the NG behaves as the K-means algorithm and only the units closer to the dataset are affected and positioned correctly. In contrast, when $\lambda_i = \lambda_f = 8.0$ (c) all NG units are well inside the dataset. Values decaying to $\lambda_f = 1.0$ (d) and $\lambda_f = 0.1$ (e) make the NG attain a better coverage of input.}
\end{figure}

\subsection{An Aside: the Role of $\lambda$ in Neural Gas} 
\label{sec:role_of_lambda} 
By design \citep{Martinetz1993}, the $\lambda$ parameter governs the level of `internal cohesion' of NG and make it less prone towards local minima in the energy function. Figure \ref{fig:decaying_lambda} gives an intuitive description of the effect that it produces. With values of both $\lambda_i$ and $\lambda_f$ close to zero, the NG behaves like the K-means algorithm: starting from the uniform, random initial configuration in Fig. \ref{fig:decaying_lambda}(a), after unsupervised training the NG attains a good coverage of the dataset (i.e. the grey area) in Fig. \ref{fig:decaying_lambda}(b) but many of its units are left unaffected and become useless. In contrast, a higher initial value $\lambda_i=8.0$ makes the NG shrink and assume the configuration shown in Fig.  \ref{fig:decaying_lambda}(c). Subsequently, the relaxation described by \eqref{eq:decaying_lambda} make the NG expand again but only within the support region of the dataset. Note that, eventually - as shown in Fig. \ref{fig:decaying_lambda}(e), all NG units are positioned inside the dataset and the coverage attained is much more effective.

The role of the $\lambda$ parameter of NG is also explored further in \citep{Parigi2015}, where it is shown experimentally its dependence on the number of units $n$ together with some interesting denoising properties that can be obtained by keeping $\lambda_f$ at relatively high values. 
   
\section{Experimental Setup}

In this section the experimental setup is described. All the experiments described were run on a workstation based on Intel(R) Xeon(R) CPU E3-1240v3 @ 3.40GHz with 16GB of RAM, Linux Ubuntu 16.04L 64bit, Python 2.7, Numpy 1.13, Scikit-Learn 0.18.

\subsection{Algorithms}

%

\subsubsection{Echo State Reservoir}

The state of the art reservoir computing model which was used as comparative reference for the experiments is Echo State Network (ESN) \citep{Jaeger2002}. In such model the transfer function is
\begin{align}
\label{eq:esn_trans_func}
	\tilde{v}_{i}(t) &:= \tanh \left( \bm{w}^{in}_i \cdot \bm{x}(t) + \bm{w}^{rec}_i \cdot \bm{v}(t-1) \right) \\
\label{eq:esn_leaky_int}
	v_i(t) &:= (1 - \gamma) v_i(t-1) + \gamma \tilde{v}_i(t)
\end{align}
In this experimental setup, values in the $\bm{W}^{in}$ matrix where drawn uniformly in the interval $[-0.5,0.5]$. Values in the $\bm{W}^{rec}$ were also drawn uniformly in the same interval. Subsequently, $\bm{W}^{rec}$ was multiplied by $\frac{r} {\rho_{rec}}$ where $r$ is a parameter and $\rho_{rec}$ is the \emph{spectral radius} of $\bm{W}^{rec}$. The justification for this procedure is described in \citep{Jaeger2002}. After what above, values in both $\bm{W}^{in}$ and $\bm{W}^{rec}$ were set to $0$ with probability $s$, where $s$ represents the \emph{sparsity} fraction of the reservoir. Note that the ESN model does not encompass a pre-training phase of any sort.

\subsubsection{Recursive Neural Gas variants}
For the purpose of performance evaluation, all the six different variants of the RNG algorithm described in Section \ref{section_2} were considered. These variants are:

\begin{enumerate}
\item \textbf{RNG-IR}, which corresponds to the original RNG account: vectors $\bm{w}^{in}$ and $\bm{w}^{rec}$ are trained in separate vector spaces;
\item \textbf{RNG-J}: vectors $\bm{w}^{in}$ and $\bm{w}^{rec}$ are trained as part of the joint vector space defined by \eqref{eq:joint_space_norm};
\item \textbf{M-RNG-IR}: RNG with masked units, vectors $\bm{w}^{in}$ and $\bm{w}^{rec}$ are trained in separate vector spaces;
\item \textbf{M-RNG-J}: RNG with masked units, vectors $\bm{w}^{in}$ and $\bm{w}^{rec}$ are trained as part of the joint vector space;
\item \textbf{M-RNG-IRM}: RNG with masked units, vectors $\bm{w}^{in}$ and $\bm{w}^{rec}$ are trained in separate vector spaces, vectors $\bm{w}^{masked}$ are updated with \eqref{eq:masked_update};
\item \textbf{M-RNG-JM}: RNG with masked units, vectors $\bm{w}^{in}$ and $\bm{w}^{rec}$ are trained as part of the joint vector space, vectors $\bm{w}^{masked}$ are updated with \eqref{eq:masked_update}.
\end{enumerate}

\subsubsection{Common readout layer}

Given that in this work we are concerned on the effects of unsupervised pre-training only, all reservoir computing systems used for the experiments were equipped with a readout layer performing a linear regression with regularization (i.e. ridge regression). The predicted output value was 
\begin{equation}
\hat{y}(t) = \bm{w}^{out} \cdot \bm{v}(t)
\end{equation}
where $\bm{w}^{out}$ is the parameter of the linear regression and $\bm{v}(t)$ is the instantaneous state of the reservoir. Actual parameters were obtained via supervised training via the equation
\begin{equation}
\label{eq:lr_optimization}
\bm{w}^{out} = (\bm{V}\bm{V}^T - \mu\bm{I})^{-1} \bm{V}^T\bm{y}
\end{equation}
where $\bm{V}:=[\bm{v}(t)]$ is the sequence of internal states of the reservoir for the training sequence, $\bm{y}:=[y(t)]$ is the sequence of true output values and $\mu$ is the regularization parameter. Eq. \eqref{eq:lr_optimization} minimizes $\bm{w}^{out}$ w.r.t. the loss function
\begin{equation}
J(\bm{y}) = \frac{1}{2} |\bm{V}\bm{w}^{out} - \bm{y}|^2 + \frac{\mu}{2}|\bm{w}^{out}|^2
\end{equation}
 
\subsection{Error measurement} 

In all experiments performed, the results were measured in terms of Normalised Root Mean Square Error (NRMSE):
\begin{equation}
\mathrm{NRMSE} :=  \sqrt{\frac{\sum_{t = 1}^{K} \left( \hat{y}(t) - y(t) \right) ^ 2}{K\,\sigma}}
\end{equation} 
where $K$ is the length of the test sequence, $y(t)$ is the true value, $\hat{y}(t)$ is the predicted value and $\sigma$ is the empirical variance in the sequence of true values. To avoid considering initialization effects, when computing RMSE values all initial pairs $(\hat{y}(t), y(t))$  were discarded up to a given $t_{washout}$ index.

\subsection{Datasets}

The proposed approach has been validated on three standard artificial benchmark datasets related to non-linear system identification and chaotic time series prediction. On those datasets, ESNs are known to perform well in prediction tasks \citep{Scardapane2016} and this makes them suitable for comparing the performances of the different reservoir computing models considered. 

\paragraph{NARMA-n: } Nonlinear autoregressive moving average dataset is a discrete-time temporal task with $n-th$ order time lag, described by the following equation in which $n$ represents the time lag, $\alpha = 0.3$, $\beta = 0.05$, $\gamma = 1.5$ and the input $x(t)$ is drawn from a uniform distribution in the interval [0,0.5] \citep{Goudarzi2014}:
\begin{equation*}
\label{eq:narma_n}
y\left(t\right) = \alpha \ y (t-1) + \beta y (t-1) \sum_{i=1}^{n} y(t-i) + \gamma \ x(t-n)  \ x(t-1) + \delta 
\end{equation*}
The value $n=10$ was used for the task.

\paragraph{Mackey-Glass time-series:} is a well-known dynamic system that, depending on the values of the parameters, displays a range of periodic and chaotic dynamics, defined by the differential equation \citep{MackeyGlass1977}:
\begin{equation*}
\dv{}{t}x(t) = b x(t) + \dfrac{a x(t - \tau)}{1 + x(t - \tau)^{10}}
\end{equation*}
which is known to produce a chaotic time series for $\tau > 16.8$. The values $a=0.2$, $b=0.1$ and $\tau=17$ were used for the task. The task is predicting the value of the sequence after a given number of steps ahead \begin{equation*}
y(t) = x(t + t_h)
\end{equation*}
where $t_h$, time horizon, is an integer constant.

\paragraph{Lorenz attractor} is another chaotic time-series \citep{Lorenz1963}, this time 3-dimensional, defined by the following set of differential equations: 
\begin{align*}
\dot{x_1} & =\sigma (x_2 - x_1) \\
\dot{x_2} & = \rho ( x_1 - x_3 ) - x_2\\
\dot{x_3} & =x_1 x_2-\beta x_3
\end{align*}
where the values $\sigma = 10$, $\rho = 28 $, $ \beta = 8/3$ were used for the task. The task is predicting the value of $\dot{x_1}$ a given number of steps ahead 
\begin{equation*}
y(t) = \dot{x_1}(t + t_h).
\end{equation*}

All sequences obtained for the above datasets were linearly rescaled in order to fit the interval $[-1, 1]$. Sequences used for training and testing were of length 10000 and 2000 respectively.

\section{Results and Discussion}

All the parameters in both RNG and ESN-based reservoir systems had been determined via a grid search. Separate procedures were applied to ESN and each RNG variant and per each dataset. In order to stress the RNG capability to adapt to different inputs, its input scaling parameter was kept fixed at $1.0$ in all training experiments.

For repeatability, each experiment was conducted with seed-controlled random sequences and each task was repeated $50$ times with different random seeds. Experiments were repeated for ESN and RNG with different number of units $n \in \{100, 200, 300, 400\}$. Further details about the experiments can be seen in the source code provided as supplementary material.

Figure \ref{fig:narma10} shows the results for the NARMA-10 task, which reveals that in this case the pre-training of RNG does not lead to advantages whatsoever. In fact, the best performance is attained by an ESN with $n=400$ yielding a NRMSE $= 0.1707, \sigma = 1.84\mathrm{e}{-4}$. In contrast, the best RNG results is obtained with a RNG-IR with $n=400$ at $NRMSE=0.7525, \sigma = 2.11\mathrm{e}{-4}$. 

Figures \ref{fig:mg17_increase_th} and \ref{fig:lambda_results} show the results of the Mackey-Glass task. In \ref{fig:mg17_increase_th} the rankings are reversed as the values of $t_h$ increase. In fact, for $t_h=10$, the ESN prevails slightly with an optimal result at NRMSE $= 0.0810, \sigma = 1.91\mathrm{e}{-4}$ vs. the result at NRMSE $= 0.0830, \sigma = 3.33\mathrm{e}{-4}$ obtained with a M-RNG-IRM. As the value of $t_h$ is increased, the performances of ESN degrade more rapidly than those of RNG, so that at $t_h=80$ the best result is obtained by M-RNG-IRM with NRMSE $= 0.14, \sigma = 1.49\mathrm{e}{-3}$ vs. the result of ESN at NRMSE $= 0.2456, \sigma = 2.84\mathrm{e}{-4}$.

Figure \ref{fig:lorenz} shows the results of the Lorenz attractor task. In this case, the pre-trained RNG is clearly prevailing with an optimal result obtained by a M-RNG-J at NRMSE $= 0.1061, \sigma = 1.19\mathrm{e}{-3}$ vs. the result of ESN at NRMSE $= 0.2837, \sigma = 2.18\mathrm{e}{-4}$.

An interesting aspect related to RNG training strategies is described in Figure \ref{fig:lambda_results}. \\Fig. \ref{fig:mg17_th20_lambda_proper} shows the results produced with values of $\lambda$ that decay according to \eqref{eq:decaying_lambda} from a very high initial value $\lambda_i=50.0$ to near zero. Such values were used for producing all the RNG results discussed so far. In contrast, Fig. \ref{fig:mg17_th20_lambda_0} shows the dramatic and negative effect of using the K-means training strategy: the performances are not good and degrade rapidly with increasing number of units, which is symptom that the RNG could organize its units effectively. The situation is only slightly better when $\lambda$ is kept at $1.0$, as in Fig. \ref{fig:mg17_th20_lambda_1}. Altogether, these figures show that, unless the appropriate training strategy is adopted, the benefit of reservoir pre-training may be lost.

Overall, the results presented show that, although RNG-based reservoir system are not prevailing in all tasks, there exists clear contexts in which reservoir pre-training is advantageous. Apart from the borderline case of the Mackey-Glass test, with the Lorenz attractor - possibly due also to the presence of multiple input variables, RNG-based reservoirs are more effective. In general, masked variants of RNG tend to perform better although this becomes more evident as the number of units increases. 

\begin{figure}[t]
\centering
    \begin{subfigure}[b]{0.48\textwidth}
        \begin{tikzpicture}
\begin{axis}[
    xlabel={\# Reservoir units},
    ylabel={NRMSE},
    xmin=0, xmax=500,
    ymin=0.0000, ymax=1,
    xtick={0, 100, 200, 300, 400},
    ytick={},
    legend pos=outer north east,
    legend cell align={left},
    ymajorgrids=true,
    grid style=dashed,
]

\addplot[
	solid,
    color=black,
    mark=*,] coordinates {
(100, 0.369823318)
(200, 0.249531996)
(300, 0.192059259)
(400, 0.170728773)
};
\addlegendentry{ESN};


\addplot[
    solid,
    color=teal,
    mark=diamond*,] coordinates {
(100, 0.709824770)
(200, 0.694272469)
(300, 0.697676882)
(400, 0.702055112)
};
\addlegendentry{RNG-IR};

\addplot[
	solid,
    color=teal,
    mark=triangle*,] coordinates {
(100, 0.752004783)
(200, 0.717679155)
(300, 0.710923043)
(400, 0.711852586)
};
\addlegendentry{RNG-J};


\addplot[
    solid, 
    color=blue,
    mark=diamond*,] coordinates {
(100, 0.708885933)
(200, 0.695620398)
(300, 0.698693459)
(400, 0.701994414)
};
\addlegendentry{M-RNG-IR};


\addplot[
    solid,
    color=blue,
    mark=triangle*,] coordinates {
(100, 0.752499912)
(200, 0.718032711)
(300, 0.711802132)
(400, 0.712989787)
};       
\addlegendentry{M-RNG-J};


\addplot[
    solid,
    color=red,
    mark=diamond*,] coordinates {
(100, 0.727927596)
(200, 0.715636939)
(300, 0.707913244)
(400, 0.704289690)
};      
\addlegendentry{M-RNG-IRM};


\addplot[
    solid,
    color=red,
    mark=triangle*,] coordinates {
(100, 0.752499911)
(200, 0.718032711)
(300, 0.711802131)
(400, 0.712989787)
};      
\addlegendentry{M-RNG-JM};

\legend{};

\end{axis}
\end{tikzpicture}
        \caption{\centering NARMA-10}
        \label{fig:narma10}
    \end{subfigure}%
    \hfill
 	\begin{subfigure}[b]{0.40\textwidth}
        \begin{tikzpicture}
\centering
\begin{axis}[%
    title={Evaluated Models},
    hide axis,
    xmin=50,
    xmax=50,
    ymin=0,
    ymax=0.4,
    legend style={draw=white!15!black, legend cell align = left, legend pos = north west}
    ]
\addlegendimage{solid, color=black, mark=*}
\addlegendentry{ESN};
\addlegendimage{solid,color=teal,mark=triangle*}
\addlegendentry{RNG-J}
\addlegendimage{solid, color=teal, mark=diamond*}
\addlegendentry{RNG-IR};
\addlegendimage{solid,color=blue,mark=diamond*}
\addlegendentry{M-RNG-IR}
\addlegendimage{solid, color=blue,mark=triangle*}
\addlegendentry{M-RNG-J}
\addlegendimage{solid,color=red,mark=diamond*}
\addlegendentry{M-RNG-IRM}
\addlegendimage{solid, color=red,mark=triangle*} 
\addlegendentry{M-RNG-JM}
\end{axis}
\end{tikzpicture}
    \end{subfigure} \\
    
	\vspace{0.2cm}
    \begin{subfigure}[b]{0.48\textwidth}
        \begin{tikzpicture}
\begin{axis}[
    title={},
    xlabel={time horizon},
    ylabel={NRMSE},
    xmin=0, xmax=90,
    ymin=0.0000, ymax=0.65,
    xtick={0, 10, 20, 30,40, 50,60,70,80},
    ytick={},
    legend pos=outer north east,
    legend cell align={left},
    ymajorgrids=true,
    grid style=dashed,
]

\addplot[
	solid,
    color=black,
    mark=*,] coordinates {
(10, 0.081022234)
(20, 0.085159831)
(40, 0.118092334)
(80, 0.245628684)
};
\addlegendentry{ESN};


\addplot[
    solid,
    color=teal,
    mark=diamond*,] coordinates {
(10, 0.097668121)
(20, 0.072068586)
(40, 0.095533903)
(80, 0.172205408)
};
\addlegendentry{RNG-IR};

\addplot[
	solid,
    color=teal,
    mark=triangle*,] coordinates {
(10, 0.098417969)
(20, 0.072545976)
(40, 0.097131851)
(80, 0.169429618)
};
\addlegendentry{RNG-J};


\addplot[
    solid, 
    color=blue,
    mark=diamond*,] coordinates {
(10, 0.096587816)
(20, 0.071837739)
(40, 0.094296318)
(80, 0.173553666)
};
\addlegendentry{M-RNG-IR};


\addplot[
    solid,
    color=blue,
    mark=triangle*,] coordinates {
(10, 0.0978438320)
(20, 0.0727264730)
(40, 0.0971221860)
(80, 0.1695735690)
};       
\addlegendentry{M-RNG-J};


\addplot[
    solid,
    color=red,
    mark=diamond*,] coordinates {
(10, 0.083029796)
(20, 0.061833267)
(40, 0.080651881)
(80, 0.14009887)
};      
\addlegendentry{M-RNG-IRM};


\addplot[
    solid,
    color=red,
    mark=triangle*,] coordinates {
(10, 0.095872805)
(20, 0.066382778)
(40, 0.086204946)
(80, 0.152248314)
};      
\addlegendentry{M-RNG-JM};

\legend{};

\end{axis}
\end{tikzpicture}
        \caption{\centering Mackey-Glass $\tau=17$, $n=300$ \newline increasing $t_h$ values}
        \label{fig:mg17_increase_th}
    \end{subfigure}%
    \hfill
   	\begin{subfigure}[b]{0.48\textwidth}
        \begin{tikzpicture}
\begin{axis}[
    xlabel={\# Reservoir units},
    ylabel={NRMSE},
    xmin=0, xmax=500,
    ymin=0.0000, ymax=0.65,
    xtick={0, 100, 200, 300, 400},
    ytick={},
    legend pos=outer north east,
    legend cell align={left},
    ymajorgrids=true,
    grid style=dashed,
]

\addplot[
	solid,
    color=black,
    mark=*,] coordinates {
(100, 0.288420028)
(200, 0.283758289)
(300, 0.295873057)
(400, 0.283730454)
};
\addlegendentry{ESN};


\addplot[
    solid,
    color=teal,
    mark=diamond*,] coordinates {
(100, 0.133963691)
(200, 0.118198629)
(300, 0.114772084)
(400, 0.114449977)
};
\addlegendentry{RNG-IR};

\addplot[
	solid,
    color=teal,
    mark=triangle*,] coordinates {
(100, 0.137560505)
(200, 0.121329968)
(300, 0.110912917)
(400, 0.106149658)
};
\addlegendentry{RNG-J};


\addplot[
    solid, 
    color=blue,
    mark=diamond*,] coordinates {
(100, 00.134689842)
(200, 00.118437)
(300, 00.115902818)
(400, 00.116125173)
};
\addlegendentry{M-RNG-IR};


\addplot[
    solid,
    color=blue,
    mark=triangle*,] coordinates {
(100, 0.136758878)
(200, 0.120372186)
(300, 0.109672172)
(400, 0.108408612)
};       
\addlegendentry{M-RNG-J};


\addplot[
    solid,
    color=red,
    mark=diamond*,] coordinates {
(100, 0.1447003710)
(200, 0.1274324280)
(300, 0.1257301970)
(400, 0.1191152310)
};      
\addlegendentry{M-RNG-IRM};


\addplot[
    solid,
    color=red,
    mark=triangle*,] coordinates {
(100, 0.143184872)
(200, 0.134039963)
(300, 0.131378724)
(400, 0.134439886)
};      
\addlegendentry{M-RNG-JM};

\legend{};

\end{axis}
\end{tikzpicture}
        \caption{\centering Lorenz attractor $t_h = 2$}
        \label{fig:lorenz}
    \end{subfigure}
\caption{}
\label{fig:main_results}
\end{figure}

\subsection{Conclusions and Future Work}

The complete description of the RNG model for pre-trainable reservoir computing system has been described and analysed, together with several proposed variants that are intended to explore different aspects of the RNG algorithm that could lead to further improvements. The results presented show the existence of benchmark tasks for which the implementation of such models seems to be clearly beneficial. Further tests in more practically-oriented application scenarios are required to assess these potential benefits in full but the results presented here, in the authors' opinion, show that this road is definitely worth investigating.  

A very interesting topic for future research is investigating the possibility of using RNG in combination with some form of Hebbian Learning, to harness its intrinsic capability to organize itself in a structure that can discover and harness possible manifold-like substructures in the state space of the system.

\begin{figure}[t]
\centering
   \begin{subfigure}[b]{0.48\textwidth}
        \begin{tikzpicture}
\begin{axis}[
    title={},
    xlabel={\# Reservoir units},
    ylabel={NRMSE},
    xmin=0, xmax=500,
    ymin=0.0000, ymax=0.65,
    xtick={0, 100, 200, 300, 400},
    ytick={},
    legend pos=outer north east,
    legend cell align={left},
    ymajorgrids=true,
    grid style=dashed,
]

\addplot[
	solid,
    color=black,
    mark=*,] coordinates {
(100, 0.1140140040)
(200, 0.0901439760)
(300, 0.0851598310)
(400, 0.080877630)
};
\addlegendentry{ESN};


\addplot[
    solid,
    color=teal,
    mark=diamond*,] coordinates {
(100, 0.076927712)
(200, 0.069884948)
(300, 0.072068586)
(400, 0.071797864)
};
\addlegendentry{RNG-IR};

\addplot[
	solid,
    color=teal,
    mark=triangle*,] coordinates {
(100, 0.102982634)
(200, 0.083394582)
(300, 0.072545976)
(400, 0.070477971)
};
\addlegendentry{RNG-J};


\addplot[
    solid, 
    color=blue,
    mark=diamond*,] coordinates {
(100, 0.07811568)
(200, 0.069606802)
(300, 0.071837739)
(400, 0.07319742)
};
\addlegendentry{M-RNG-IR};


\addplot[
    solid,
    color=blue,
    mark=triangle*,] coordinates {
(100, 0.102969965)
(200, 0.083484257)
(300, 0.072726473)
(400, 0.070232675)
};       
\addlegendentry{M-RNG-J};


\addplot[
    solid,
    color=red,
    mark=diamond*,] coordinates {
(100, 0.0726759180)
(200, 0.0653809210)
(300, 0.0618332670)
(400, 0.0616496660)
};      
\addlegendentry{M-RNG-IRM};


\addplot[
    solid,
    color=red,
    mark=triangle*,] coordinates {
(100, 0.091333799)
(200, 0.071818943)
(300, 0.066382778)
(400, 0.06496275)
};      
\addlegendentry{M-RNG-JM};

\legend{};

\end{axis}
\end{tikzpicture}
        \caption{\centering Mackey-Glass $\tau=17, t_h=20$ \newline RNG $\lambda_i=50.0, \lambda_f=0.01$}
       	\label{fig:mg17_th20_lambda_proper}        
    \end{subfigure}%
    \hfill
 	\begin{subfigure}[b]{0.40\textwidth}
        \begin{tikzpicture}
\centering
\begin{axis}[%
    title={Evaluated Models},
    hide axis,
    xmin=50,
    xmax=50,
    ymin=0,
    ymax=0.4,
    legend style={draw=white!15!black, legend cell align = left, legend pos = north west}
    ]
\addlegendimage{solid, color=black, mark=*}
\addlegendentry{ESN};
\addlegendimage{solid,color=teal,mark=triangle*}
\addlegendentry{RNG-J}
\addlegendimage{solid, color=teal, mark=diamond*}
\addlegendentry{RNG-IR};
\addlegendimage{solid,color=blue,mark=diamond*}
\addlegendentry{M-RNG-IR}
\addlegendimage{solid, color=blue,mark=triangle*}
\addlegendentry{M-RNG-J}
\addlegendimage{solid,color=red,mark=diamond*}
\addlegendentry{M-RNG-IRM}
\addlegendimage{solid, color=red,mark=triangle*} 
\addlegendentry{M-RNG-JM}
\end{axis}
\end{tikzpicture}
    \end{subfigure} \\
    
	\vspace{0.3cm}
   	\begin{subfigure}[b]{0.48\textwidth}
        \begin{tikzpicture}
\begin{axis}[
    title={},
    xlabel={\# Reservoir units},
    ylabel={NRMSE},
    xmin=0, xmax=500,
    ymin=0.0000, ymax=0.65,
    xtick={0, 100, 200, 300, 400},
    ytick={},
    legend pos=outer north east,
    legend cell align={left},
    ymajorgrids=true,
    grid style=dashed,
]

\addplot[
	solid,
    color=black,
    mark=*,] coordinates {
(100, 0.114014004)
(200, 0.090143976)
(300, 0.085159831)
(400, 0.08087763)
};
\addlegendentry{ESN};


\addplot[
    solid,
    color=teal,
    mark=diamond*,] coordinates {
(100, 0.157645779)
(200, 0.245291892)
(300, 0.352558969)
(400, 0.601300895)
};
\addlegendentry{RNG-IR};

\addplot[
	solid,
    color=teal,
    mark=triangle*,] coordinates {
(100, 0.123066003)
(200, 0.236155503)
(300, 0.332424925)
(400, 0.433366268)
};
\addlegendentry{RNG-J};


\addplot[
    solid, 
    color=blue,
    mark=diamond*,] coordinates {
(100, 0.181834598)
(200, 0.246479443)
(300, 0.380773587)
(400, 0.601576303)
};
\addlegendentry{M-RNG-IR};


\addplot[
    solid,
    color=blue,
    mark=triangle*,] coordinates {
(100, 0.137528431)
(200, 0.246609688)
(300, 0.337376975)
(400, 0.444534104)
};       
\addlegendentry{M-RNG-J};


\addplot[
    solid,
    color=red,
    mark=diamond*,] coordinates {
(100, 0.138318307)
(200, 0.237562861)
(300, 0.328548798)
(400, 0.436875964)
};      
\addlegendentry{M-RNG-IRM};


\addplot[
    solid,
    color=red,
    mark=triangle*,] coordinates {
(100, 0.192925566)
(200, 0.249126827)
(300, 0.384136415)
(400, 0.591302719)
};      
\addlegendentry{M-RNG-JM};

\legend{};

\end{axis}
\end{tikzpicture}
		\caption{\centering Mackey-Glass $\tau=17, t_h=20$ \newline RNG $\lambda_i=\lambda_f=0$}
		\label{fig:mg17_th20_lambda_0}
    \end{subfigure}
    \hfill
   \begin{subfigure}[b]{0.48\textwidth}
        \begin{tikzpicture}
\begin{axis}[
    title={},
    xlabel={\# Reservoir units},
    ylabel={NRMSE},
    xmin=0, xmax=500,
    ymin=0.0000, ymax=0.65,
    xtick={0, 100, 200, 300, 400},
    ytick={},
    legend pos=outer north east,
    legend cell align={left},
    ymajorgrids=true,
    grid style=dashed,
]

\addplot[
    solid,
    color=black,
    mark=*,] coordinates {
(100, 0.114014004)
(200, 0.090143976)
(300, 0.085159831)
(400, 0.08087763)
};
\addlegendentry{ESN};


\addplot[
    solid,
    color=teal,
    mark=diamond*,] coordinates {
(100, 0.134026133)
(200, 0.220795602)
(300, 0.304770883)
(400, 0.380808645)
};
\addlegendentry{RNG-IR};

\addplot[
    solid,
    color=teal,
    mark=triangle*,] coordinates {
(100, 0.101208967)
(200, 0.17732412)
(300, 0.240998563)
(400, 0.252371558)
};
\addlegendentry{RNG-J};


\addplot[
    solid, 
    color=blue,
    mark=diamond*,] coordinates {
(100, 0.156348667)
(200, 0.226602692)
(300, 0.340079893)
(400, 0.434364291)
};
\addlegendentry{M-RNG-IR};


\addplot[
    solid,
    color=blue,
    mark=triangle*,] coordinates {
(100, 0.122777016)
(200, 0.18528993)
(300, 0.241475987)
(400, 0.253610768)
};       
\addlegendentry{M-RNG-J};


\addplot[
    solid,
    color=red,
    mark=diamond*,] coordinates {
(100, 0.158031934)
(200, 0.221278357)
(300, 0.301671217)
(400, 0.345845634)
};      
\addlegendentry{M-RNG-IRM};


\addplot[
    solid,
    color=red,
    mark=triangle*,] coordinates {
(100, 0.095999316)
(200, 0.139229999)
(300, 0.16728577)
(400, 0.17706522)
};      
\addlegendentry{M-RNG-JM};

\legend{};

\end{axis}
\end{tikzpicture}
		\caption{\centering Mackey-Glass $\tau=17, t_h=20$ \newline RNG $\lambda_i=\lambda_f=1.0$}
		\label{fig:mg17_th20_lambda_1}
    \end{subfigure}%
\caption{}
\label{fig:lambda_results}
\end{figure}

\onecolumn

\bibliographystyle{plain}
\bibliography{bibliography/trainable-reservoir-computing-with-recursive-neural-gas}
\end{document}